\documentclass[letterpaper, 10 pt, conference]{ieeeconf}

\IEEEoverridecommandlockouts

\usepackage{graphics}
\usepackage{epsfig}
\usepackage{mathptmx}
\usepackage{times}
\usepackage{amsmath}
\usepackage{amssymb}
\usepackage{url}
\usepackage[dvipsnames]{xcolor}

\title{\LARGE \bf
SplatLabel: Pseudo-Labelling through 4D Gaussian Splatting
}

\author{}
\author{Nitya Nanvani$^{1,2}$ \qquad\qquad Andras Palffy$^{1}$ \qquad\qquad Holger Caesar$^{2}$%
\thanks{$^{1}$Authors are with Perciv AI, The Netherlands}%
\thanks{$^{2}$Authors are with Delft University of Technology, The Netherlands}
}

\begin{document}

\makeatletter
\let\@oldmaketitle\@maketitle
\renewcommand{\@maketitle}{\@oldmaketitle
  \vspace{0.5em}
  \begin{center}
    \includegraphics[width=0.95\textwidth]{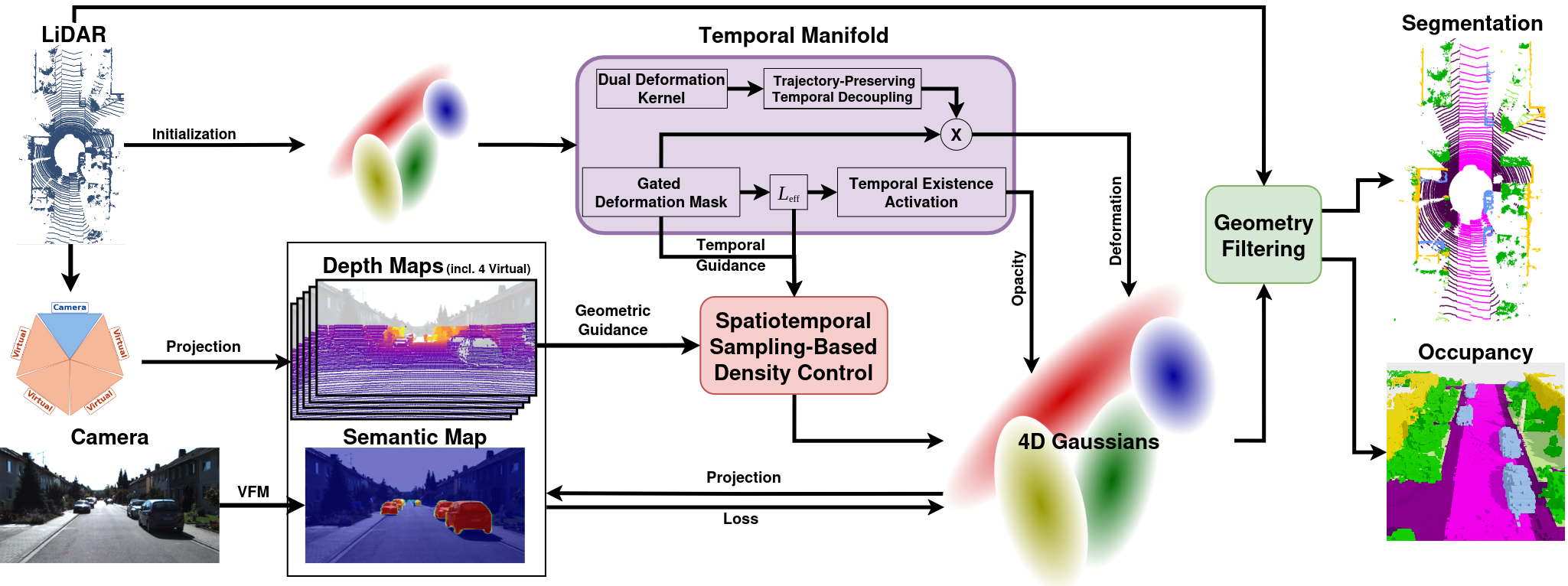}
    \def\@captype{figure}
    \caption{\textbf{SplatLabel Architecture Overview.} SplatLabel optimizes a 4D Gaussian representation governed by an explicit temporal manifold. \textbf{(Left)} 2D images and LiDAR point clouds are distilled to structurally ground the scene. \textbf{(Middle)} Dynamics are explicitly modeled by parameterizing the trajectories and lifespans of individual primitives. \textbf{(Right)} The unified representation yields high-confidence 3D pseudo-labels for multiple downstream tasks.}
    \label{fig:method}
  \end{center}
  \vspace{-1.5em}
}
\makeatother

\bstctlcite{IEEEexample:BSTcontrol}

\maketitle
\thispagestyle{empty}
\pagestyle{empty}

\setcounter{figure}{1}

\begin{abstract}
While 2D Vision Foundation Models offer a pathway to automate 3D semantic pseudo-labelling, translating these priors into robust 3D representations typically requires complex heuristics or multi-model ensembles. We introduce SplatLabel, an automated pipeline that leverages a 4D Gaussian representation to extract LiDAR segmentation with predictive confidence, as well as semantic occupancy grids at arbitrary voxel resolutions. At its core, SplatLabel handles dynamic environments through an explicit temporal manifold that models the trajectories and lifespans of individual 3D primitives. This allows the system to accurately track moving actors and strictly define when objects appear and disappear, completely eliminating the need for pre-annotated 3D bounding boxes. To robustly support this dynamic tracking, the representation is grounded by structural and semantic priors: we guide scene geometry in unobserved regions by integrating 360-degree LiDAR via virtual depth maps, and rather than relying on domain-specific prompt engineering, we directly distill continuous soft probabilities from 2D models to inherently resolve semantic ambiguities over time and space. Finally, to accurately reflect the real-world trade-off between precision and recall, we reframe pseudo-label evaluation as a selective classification task using a generalized risk-recall metric. Experiments on SemanticKITTI demonstrate that SplatLabel consistently outperforms state-of-the-art baselines across multiple recall levels, establishing a highly robust framework for both 3D LiDAR segmentation and occupancy prediction.
\end{abstract}

\begin{figure*}[ht]
    \centering
    \includegraphics[width=\textwidth]{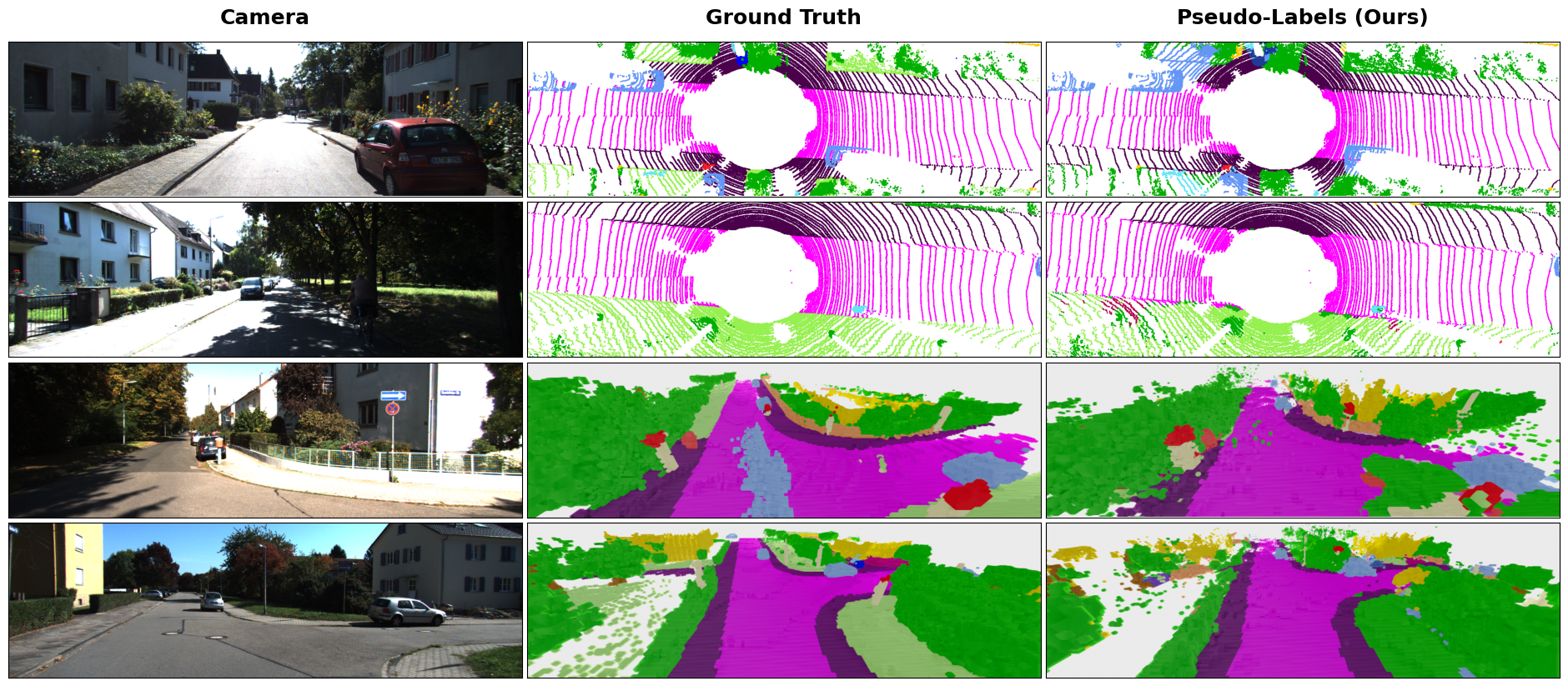}
    \caption{Qualitative Results on SemanticKITTI~\cite{behley_semantickitti_2019}. \textbf{Top two rows:} Bird's-Eye View (BEV) LiDAR segmentation. \textbf{Bottom two rows:} Semantic occupancy grids. \textbf{Left:} Reference RGB images. \textbf{Middle:} Ground truth annotations. \textbf{Right:} Our complete 360-degree predictions. Crucially, our predictions encompass all points outside the camera's Field of View (FOV). Furthermore, our explicit temporal tracking resolves the artifactual moving-object ``trails'' present in the occupancy ground truth (visible in row 3), cleanly separating dynamic actors from the static background.}
    \label{fig:qualitative_seg}
\end{figure*}

\section{Introduction}
\label{sec:intro}

Comprehensive 3D semantic perception is a fundamental prerequisite for autonomous systems~\cite{behley_semantickitti_2019, caesar_nuscenes_2020}. However, the manual annotation required for training robust models is prohibitively expensive, prompting the use of 2D Vision Foundation Models (VFMs) to automatically generate 3D pseudo-labels~\cite{peng_openscene_2023, gebraad_leap_2025, zhou_autoocc_2025, ghilotti_unilips_2026}. Concurrently, 3D Gaussian Splatting (3DGS) has emerged as a powerful explicit radiance field representation~\cite{kerbl_3d_2023}. Yet, despite its rapid adoption, 3DGS optimization remains overwhelmingly focused on novel view synthesis, leaving its potential as a continuous backbone for spatial pseudo-labelling underutilized.

Current automated 3D pseudo-labelling pipelines face critical bottlenecks when moving beyond static environments. Most notably, modeling dynamic scenes typically requires pre-annotated 3D bounding boxes to cleanly separate moving actors from static backgrounds~\cite{zhou_drivinggaussian_2024, chen_omnire_2025}, creating a circular dependency that defeats the purpose of annotation-free pseudo-labelling. Furthermore, extracting consistent semantic features and integrating 360-degree LiDAR geometry to support these scenes demands complex heuristics, multi-model ensembles, or computationally heavy spherical rasterization to counteract temporal inconsistencies and visual occlusion~\cite{gebraad_leap_2025, ghilotti_unilips_2026, giacomini_splat-loam_2025, kung_lihi-gs_2025}. Finally, existing baselines evaluate pseudo-label quality strictly on successfully labeled points, fundamentally ignoring the crucial trade-off between arbitrary recall levels and overall label accuracy.

To solve these challenges, we propose SplatLabel, an automated 4D Gaussian Splatting pipeline designed specifically for 3D semantic pseudo-labelling and volumetric occupancy prediction. Rather than treating motion as a post-processing step or relying on implicit global deformations, SplatLabel unifies explicit spatial representation, structural densification, and kinematic tracking into a single continuous optimization framework driven by a fully parameterized temporal manifold. By doing so, we address the core limitations of existing pipelines and establish a robust, annotation-free foundation for dynamic scene understanding. Our main contributions are summarized as follows:

\noindent \textbf{Explicit Temporal Manifold for Dynamic Scenes:} We parameterize motion and temporal existence boundaries as intrinsic properties of individual 3D Gaussians. By utilizing a Dual Deformation Kernel~\cite{lin_gaussian-flow_2024} coupled with a Gated Deformation Mask and explicit temporal lifespans, SplatLabel achieves robust, object-centric tracking. This naturally separates moving actors from the static background across time without the need for pre-annotated 3D bounding boxes~\cite{zhou_drivinggaussian_2024, chen_omnire_2025} or unpredictable black-box motion models~\cite{song_coda-4dgs_2025}.

\noindent \textbf{Spatiotemporal Distillation and Geometric Priors:} We fundamentally advance 3DGS optimization by dynamically regulating primitive capacity through LiDAR priors. By backpropagating continuous soft semantics directly from 2D VFMs~\cite{carion_sam_2026}, our continuous optimization inherently filters out single-frame noise and resolves categorical ambiguities through spatiotemporal consensus, eliminating the need for domain-specific prompt engineering~\cite{gebraad_leap_2025, ghilotti_unilips_2026}.

\noindent \textbf{Pseudo-Labelling as Selective Classification:} We establish a mathematically grounded framework to evaluate pseudo-labelling systems. By reformulating evaluation as a selective classification task using the class-averaged Area Under the Generalized Risk-Coverage (AUGRC)~\cite{traub_overcoming_2024} metric, we directly capture the risk of injecting noisy labels into downstream training, revealing the true performance-recall trade-offs ignored by standard baselines~\cite{gebraad_leap_2025, ghilotti_unilips_2026}.
\section{Related Work}
\label{sec:related_work}

\subsection{Spatiotemporal Semantic Distillation in 3DGS}
3DGS models scenes using an explicit radiance field of anisotropic Gaussian primitives~\cite{kerbl_3d_2023}. Typically initialized from Structure-from-Motion (SfM) point clouds~\cite{schonberger_structure--motion_2016}, each discrete primitive is parameterized by a 3D center, covariance, opacity, and spherical harmonics~\cite{kerbl_3d_2023}. While 3DGS achieves highly efficient rendering, its original photometric gradient-based densification heuristics are extremely fragile~\cite{kerbl_3d_2023, kheradmand_3d_2024}. Crucially, introducing semantic and depth supervision fundamentally alters these loss scales, which vary significantly depending on the total number of classes or the specific foundation model utilized, causing standard densification heuristics to struggle. To address this, recent work reformulates 3DGS optimization as a Markov Chain Monte Carlo (MCMC) sampling process via Stochastic Gradient Langevin Dynamics (SGLD)~\cite{kheradmand_3d_2024, brosse_promises_2018}. This approach naturally explores the scene landscape and manages primitive capacity irrespective of disrupted gradient scales~\cite{kheradmand_3d_2024}.

Beyond geometry, Gaussians explicitly store high-dimensional semantic features without structural modifications~\cite{zhu_3d_2024, kerbl_3d_2023}, enabling memory-efficient semantic occupancy prediction~\cite{huang_gaussianformer_2025, pavkovic_gaussianfusionocc_2025} and sensor fusion~\cite{bai_rags_2026, montiel-marin_gaussiancar_2026}. Concurrently, 3D semantic pseudo-labelling has advanced by transferring knowledge from 2D VFMs to 3D spaces~\cite{peng_openscene_2023, gebraad_leap_2025, zhou_autoocc_2025, ghilotti_unilips_2026}. However, extracting consistent semantics typically demands complex prompt formulation or multi-model ensembles to counteract temporal inconsistencies and single-frame label flickering~\cite{gebraad_leap_2025, ghilotti_unilips_2026}. SplatLabel overcomes this by directly distilling continuous soft logits into a 4D Gaussian representation, natively converging on a unified semantic identity across time and space.

\subsection{LiDAR-Guided Geometric Optimization}
While VFMs provide rich 2D semantic priors, accurately lifting these labels into 3D requires robust underlying geometry. The integration of LiDAR into 3DGS typically enforces structural constraints through pseudo-depth projection onto camera planes~\cite{yan_street_2025, huang_textits3gaussian_2024}, but frequently discards LiDAR measurements outside the visual frustum~\cite{hess_splatad_2025, kung_lihi-gs_2025, yan_street_2025}. To utilize 360-degree sweeps, recent methods develop explicit LiDAR sensor models utilizing specialized spherical rasterization~\cite{hess_splatad_2025, kung_lihi-gs_2025}, which introduce significant computational complexity. Crucially, prior approaches treat LiDAR strictly as a post-rendering supervision signal rather than an explicit densification trigger. To fully utilize 360-degree sweeps without heavy spherical rasterizers, SplatLabel introduces Virtual Depth Maps and a Geometry-Guided sampler that actively spawns primitives in structurally inconsistent regions.

\subsection{Dynamic Scene Modeling and the Temporal Manifold}
Despite the geometric fidelity achieved in static scenes, real-world environments are inherently dynamic. Importantly, the vast majority of continuous 4DGS frameworks are designed exclusively for novel view synthesis~\cite{zhou_drivinggaussian_2024, chen_omnire_2025, lin_gaussian-flow_2024, song_coda-4dgs_2025, yan_street_2025, hess_splatad_2025, li_spacetime_2024, chen_periodic_2026}, largely overlooking how these rich spatiotemporal representations can be repurposed for dense scene annotation.

Supervised frameworks achieve high geometric fidelity by separating moving foregrounds from static backgrounds~\cite{zhou_drivinggaussian_2024, chen_omnire_2025}. Annotation-free approaches utilize either implicit global environmental deformations~\cite{wu_4d_2024, song_coda-4dgs_2025}, or explicit representations that treat motion as an intrinsic property of the 3D primitives~\cite{chen_periodic_2026, lin_gaussian-flow_2024, li_spacetime_2024}. However, while implicit methods like CODA-4DGS~\cite{song_coda-4dgs_2025} effectively model dynamic context, they allow semantic features to deform and alter across time. This continuous semantic morphing contradicts physical reality for label extraction, where an object's core semantic identity must remain firmly grounded and constant. Conversely, while explicit models provide the object-centric tracking required for pseudo-labelling, isolating dynamic actors typically requires pre-annotated 3D bounding boxes, creating a circular dependency for self-supervised systems. SplatLabel resolves this through a bounding-box-free explicit Dual Deformation Kernel~\cite{lin_gaussian-flow_2024} and a fully parameterized temporal manifold, cleanly separating moving actors from static backgrounds while strictly preserving their semantic identities.

\subsection{Selective Classification for Pseudo-Label Evaluation}
Even with robust explicit tracking, self-supervised pipelines inevitably generate noisy predictions. Consequently, evaluating pseudo-label quality is uniquely challenging. Existing baselines conventionally measure accuracy only on the fraction of points they manage to annotate, failing to account for the widely varying and arbitrary recall levels across different methods. To address this, we draw from the broader machine learning literature, where selective classification is utilized to explicitly evaluate the trade-off between coverage (recall) and accuracy~\cite{el-yaniv_foundations_2010, geifman_selective_2017}. However, simply adopting standard selective metrics like the Area Under the Risk-Coverage curve (AURC)~\cite{geifman_selective_2017} introduces new issues: they calculate risk strictly over accepted predictions, which disproportionately penalizes high-confidence failures~\cite{traub_overcoming_2024} and frequently mask bias by disproportionately rejecting minority classes under severe dataset imbalance~\cite{saglam_selective_2026}. To overcome both the limitations of standard pseudo-label evaluation and the flaws of naive selective metrics, SplatLabel reformulates LiDAR pseudo-labelling as a selective classification task utilizing a class-averaged Area Under the Generalized Risk-Coverage (AUGRC) metric~\cite{traub_overcoming_2024}, providing a holistic, bias-aware measure of label reliability.
\section{Methodology}
\label{sec:methodology}

\subsection{Problem Formulation and Overview}
Given a continuous sequence of LiDAR point clouds and 2D images, our objective is to generate dense 3D semantic pseudo-labels and volumetric semantic occupancy grids. The fundamental challenge in continuous dynamic scene reconstruction is accurately tracking moving actors while ensuring the static environment remains stable over time. 

To resolve this, we formulate the scene as a continuous 4D representation strictly governed by an explicit temporal manifold. As illustrated in Figure \ref{fig:method}, our pipeline centers on this manifold by treating motion and temporal lifespan as intrinsic properties of individual Gaussian primitives. While multimodal inputs (LiDAR and 2D semantics) are distilled to manage primitive capacity, the core of our approach lies in explicit kinematic tracking and temporal existence parameterization. This formulation allows the system to accurately define when and where objects appear, seamlessly separating dynamic actors from static backgrounds to achieve a globally consistent 4D reconstruction.

\begin{figure}[t]
  \centering
   \includegraphics[width=\linewidth]{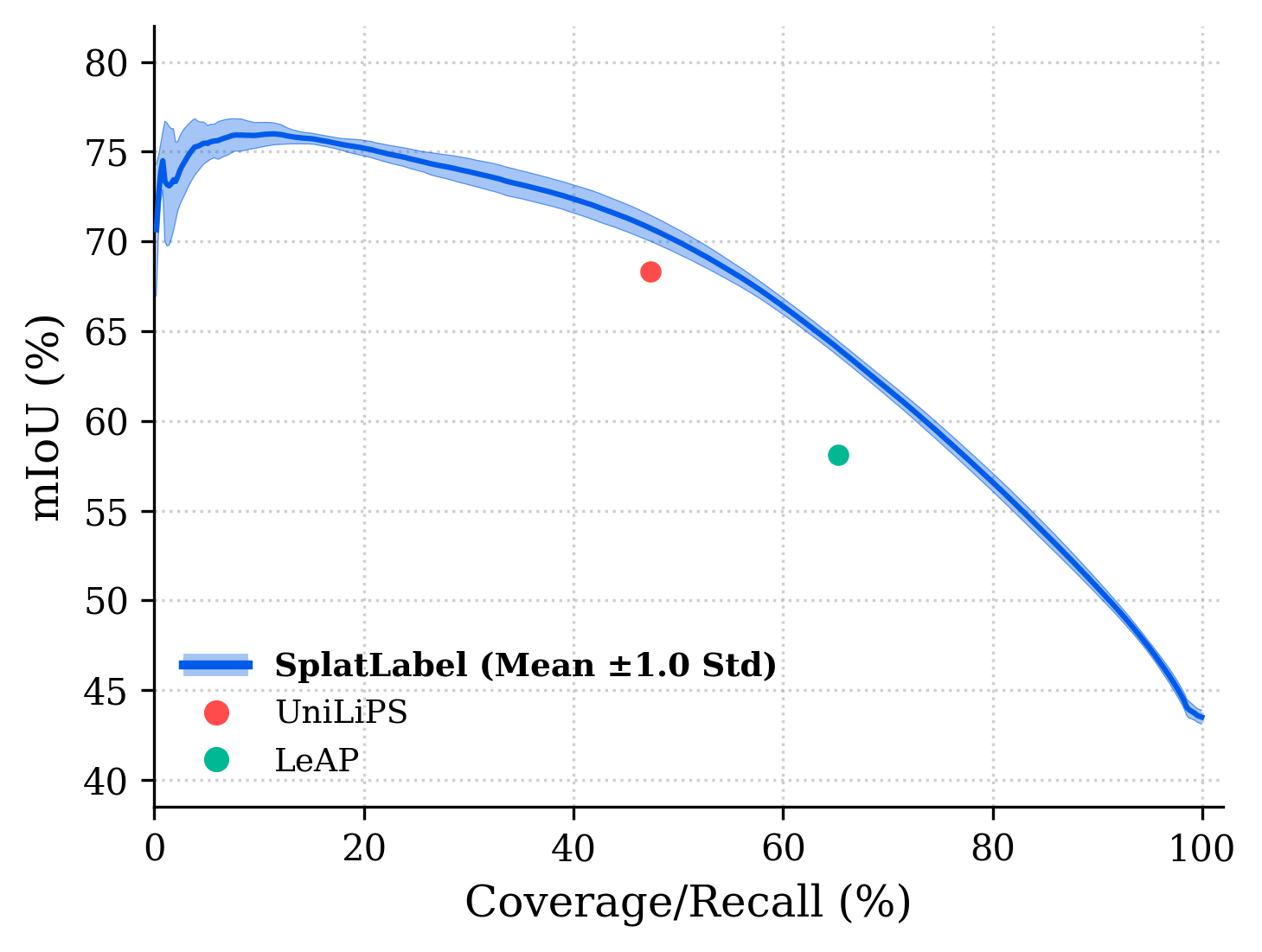}
   \caption{\textbf{Performance-recall trade-off on SemanticKITTI~\cite{behley_semantickitti_2019}.} SplatLabel provides a continuous performance-recall profile, displaying the mean mIoU and standard deviation across five random seeds. Our method's tight variance and consistent upper bound strictly dominate the single-operating-point baselines, LeAP~\cite{gebraad_leap_2025} and UniLiPs~\cite{ghilotti_unilips_2026}.}
   \label{fig:coverage_tradeoff}
\end{figure}

\subsection{Spatiotemporal Distillation and Geometric Priors}

To build a robust foundation for our temporal manifold, we must first accurately ground our Gaussians in both semantic and geometric reality. Standard photometric gradient-based densification heuristics fail under the altered loss scales introduced by semantic and depth supervision. Therefore, we abandon fragile densification heuristics and build upon stochastic state transitions (MCMC via SGLD)~\cite{kheradmand_3d_2024} to dynamically regulate primitive capacity. We extend this framework to natively incorporate multimodal 2D and 3D priors.

\subsubsection{Continuous Semantic Distillation}
To overcome domain-specific prompt limitations, we directly extract raw logits from Segment Anything Model (SAM) 3~\cite{carion_sam_2026} by querying base dataset labels. This bypasses complex prompt manipulation, yielding dense, continuous soft labels. By backpropagating the semantic loss directly into the Gaussian parameters and aggregating gradients across the entire spatiotemporal sequence, our continuous optimization intrinsically smooths frame-to-frame label flickering and resolves categorical ambiguities (e.g., `car' versus `vehicle').

\subsubsection{Omnidirectional Geometric Supervision} 
Standard camera-frame depth supervision discards crucial 360-degree LiDAR geometry. To supervise visually occluded regions without abandoning the efficient pinhole rasterizer, we introduce \textit{Virtual Depth Maps}. By explicitly placing artificial cameras in unobserved areas, we project the full LiDAR point cloud to generate virtual depth targets. Furthermore, we integrate these raw LiDAR depth cues directly into our density controller as structural sampling priors. When the system detects regions exhibiting high structural inconsistency between the Gaussian distribution and LiDAR data, it actively prioritizes spawning and adjusting primitives directly within those areas.

\subsection{Temporal Manifold Parameterization}
To accurately model dynamic environments without decoupling moving actors from global environmental deformations, we parameterize motion and temporal existence as an explicit temporal manifold intrinsic to individual Gaussian primitives. This unifies spatial tracking and temporal boundaries into a single continuous representation.

\subsubsection{Dual Deformation Kernel}
Trajectory changes are captured via a Dual Deformation Kernel~\cite{lin_gaussian-flow_2024}, combining an $n$-degree polynomial framework with an $n$-degree Fourier noise network to model both smooth linear paths and complex, high-frequency variations:
\begin{equation}
\label{eq:dual_kernel}
\resizebox{0.89\hsize}{!}{$
\displaystyle \Delta(t) = \sum_{k=1}^n \alpha_k t^k + \sum_{m=1}^n \beta_m \sin(2 \pi m t) + \gamma_m \cos(2 \pi m t)
$}
\end{equation}

\subsubsection{Gated Deformation Mask}
Uniformly applying explicit deformation destabilizes static backgrounds. To isolate static geometry, we introduce a learned Sigmoid-Gated Deformation Mask~\cite{song_coda-4dgs_2025}, which clamps deformation kernels to zero for background structures. Furthermore, moving objects with poorly converged trajectories often suffer low opacity and are consequently pruned. To address this, we integrate the mask directly into our sampling prior as a dynamic-weighted relocalization mechanism. By scaling primitive respawn probabilities with this motion mask, the state transition backend actively prioritizes sampling along dynamic trajectories, ensuring robust reconstruction of moving actors.

\subsubsection{Deformation-Coupled Lifespan}
To enforce distinct temporal windows for moving actors while preserving static backgrounds indefinitely, we assign each Gaussian a base lifespan ($L_{\text{base}}$) and couple it to the Gated Deformation Mask ($M_{\text{deform}}$) to compute an effective lifespan ($L_{\text{eff}}$):
\begin{equation}
L_{\text{eff}} = (1 - M_{\text{deform}}) + M_{\text{deform}} \cdot L_{\text{base}}
\label{eq:effective_lifespan}
\end{equation}
This formulation ensures that static primitives ($M_{\text{deform}} \to 0$) retain a full, sequence-wide footprint of $1.0$, while dynamic elements ($M_{\text{deform}} \to 1$) are tightly bounded by their localized lifespan. Primitives whose $L_{\text{eff}}$ falls below a minimum temporal threshold are actively culled and respawned, preventing the waste of computational capacity on transient artifacts.

\subsubsection{Temporal Existence Activation}
To bridge continuous gradient optimization and discrete physical boundaries, we introduce the Temporal Existence Activation:
\begin{equation}
\alpha_{t} = \text{sigmoid}\left(k \cdot \left(1 - \frac{|t - \tau|}{L_{\text{eff}}}\right)\right)
\end{equation}
This activation value acts as a multiplier to the Gaussian's base spatial opacity. It is evaluated symmetrically around the primitive's learned temporal center ($\tau \in [0, 1]$) within its effective lifespan $L_{\text{eff}}$, where $t$ represents the normalized global sequence time. Crucially, $k$ acts as a progressive sharpness parameter. During initial training, a low $k$ creates a smooth distribution for free temporal gradient propagation. As optimization converges, $k$ is systematically scaled upward, sharpening into a rigid box-car distribution that strictly enforces temporal existence boundaries.

\subsubsection{Trajectory-Preserving Temporal Decoupling}
To capture highly complex, non-linear motions, the system performs temporal exploration (perturbing $\tau$ to discover optimal temporal alignments) and spatiotemporal cloning (splitting a dynamic primitive's lifespan into sequential segments to accurately represent piecewise motion). However, actively shifting this reference center within a localized kernel fundamentally misaligns the learned polynomial and Fourier coefficients of Equation \ref{eq:dual_kernel}. 

To enable these temporal shifts without corrupting trajectory convergence, we propose a decoupled, global-time based deformation kernel, $\Delta_{\text{eff}}(t, \tau)$:
\begin{equation}
\Delta_{\text{eff}}(t, \tau) = \Delta(t) - \Delta(\tau)
\end{equation}
Under this parameterization, the underlying trajectory $\Delta(t)$ operates strictly on global sequence time. When the system updates $\tau$ to a new value $\tau'$, the Gaussian's spatial parameters are exactly and efficiently compensated by applying the analytic offset $\Delta(\tau') - \Delta(\tau)$ to the base reference position, guaranteeing that temporal exploration and cloning can freely optimize the existence window without destroying established kinematics.

% Color definitions for all tables
\definecolor{cCar}{RGB}{100, 150, 245}
\definecolor{cBicycle}{RGB}{100, 230, 245}
\definecolor{cMotorcycle}{RGB}{30, 60, 150}
\definecolor{cOtherVehicle}{RGB}{0, 0, 255}
\definecolor{cPerson}{RGB}{255, 30, 30}
\definecolor{cRoad}{RGB}{255, 0, 255}
\definecolor{cSidewalk}{RGB}{75, 0, 75}
\definecolor{cOtherGround}{RGB}{175, 0, 75}
\definecolor{cManmade}{RGB}{255, 200, 0}
\definecolor{cVegetation}{RGB}{0, 175, 0}
\definecolor{cTerrain}{RGB}{150, 240, 80}
\definecolor{cPole}{RGB}{255, 240, 150}

\begin{table*}[t]
\centering
\footnotesize
\caption{\textbf{Detailed LiDAR Segmentation Performance.} Comprehensive per-class IoU evaluated on the SemanticKITTI validation set~\cite{behley_semantickitti_2019} at the specific recall levels of baseline methods. Best results are highlighted in \textbf{bold}.}
\label{tab:supp_comparison}
\renewcommand{\arraystretch}{1.2}
\setlength{\tabcolsep}{3.5pt}
\begin{tabular}{l | c c | c c c c c c c c c c c | l | c c}
\hline
\multicolumn{14}{c|}{\textbf{Recall: 65.3\%}} & \multicolumn{3}{c}{\textbf{Recall: 47.32\%}} \\
\textbf{Method} & 
\rotatebox{90}{mIoU \% $\uparrow$} & 
\rotatebox{90}{cat. mIoU \% $\uparrow$} & 
\rotatebox{90}{\textcolor{cCar}{\rule{1.5ex}{1.5ex}}\hspace{4pt}Car} & 
\rotatebox{90}{\textcolor{cBicycle}{\rule{1.5ex}{1.5ex}}\hspace{4pt}Bicycle} & 
\rotatebox{90}{\textcolor{cMotorcycle}{\rule{1.5ex}{1.5ex}}\hspace{4pt}Motorcycle} & 
\rotatebox{90}{\textcolor{cOtherVehicle}{\rule{1.5ex}{1.5ex}}\hspace{4pt}Oth.-vehicle} & 
\rotatebox{90}{\textcolor{cPerson}{\rule{1.5ex}{1.5ex}}\hspace{4pt}Person} & 
\rotatebox{90}{\textcolor{cRoad}{\rule{1.5ex}{1.5ex}}\hspace{4pt}Road} & 
\rotatebox{90}{\textcolor{cSidewalk}{\rule{1.5ex}{1.5ex}}\hspace{4pt}Sidewalk} & 
\rotatebox{90}{\textcolor{cOtherGround}{\rule{1.5ex}{1.5ex}}\hspace{4pt}Oth.-ground} & 
\rotatebox{90}{\textcolor{cManmade}{\rule{1.5ex}{1.5ex}}\hspace{4pt}Manmade} & 
\rotatebox{90}{\textcolor{cVegetation}{\rule{1.5ex}{1.5ex}}\hspace{4pt}Vegetation} & 
\rotatebox{90}{\textcolor{cTerrain}{\rule{1.5ex}{1.5ex}}\hspace{4pt}Terrain} & 
\textbf{Method} & 
\rotatebox{90}{mIoU \% $\uparrow$} & 
\rotatebox{90}{cat. mIoU \% $\uparrow$} \\
\hline
\small
LeAP~\cite{gebraad_leap_2025} & 58.1 & 81.6 & \textbf{92.5} & 24.5 & 26.8 & 27.3 & \textbf{71.7} & 93.9 & 73.7 & 0.0 & 69.4 & \textbf{82.9} & \textbf{76.2} & UniLiPs~\cite{ghilotti_unilips_2026} & 68.3 & 86.6 \\
\textbf{SplatLabel (Ours)} & \textbf{64.2} & \textbf{82.1} & 84.3 & \textbf{45.7} & \textbf{71.9} & \textbf{84.6} & 70.7 & \textbf{96.2} & \textbf{84.3} & \textbf{10.0} & \textbf{79.3} & 71.9 & 7.4 & \textbf{SplatLabel (Ours)} & \textbf{71.6} & \textbf{87.6} \\
\hline
\end{tabular}
\end{table*}

\subsection{Deformation and Trajectory Regularization}
While the explicit temporal manifold effectively captures 4D kinematics, it requires regularization to converge to physically plausible states. To ensure the Gated Deformation Mask functions as a strict binary classifier rather than a continuous scaling factor, we apply a combined $\mathcal{L}_1$ sparsity penalty and a self-entropy minimization loss. This exerts downward pressure on static geometry while driving mask values strictly toward bistable states ($0$ or $1$). 
Simultaneously, to prevent overfitting in the polynomial motion framework, we apply a small $\mathcal{L}_1$ penalty directly to the trajectory coefficients, which inherently favors smooth paths unless complex motion is heavily supported by structural gradients. \looseness = -1

\subsection{Unified Downstream Extraction}
To demonstrate the versatility of our continuous 4D representation, we extract discrete pseudo-labels for two distinct downstream tasks: LiDAR point cloud segmentation and semantic occupancy prediction. Because 3DGS inherently prioritizes 2D camera coverage, it often generates artifacts in regions lacking depth priors. To ensure high fidelity during this continuous-to-discrete translation, we first apply a Voxel-Based Geometric Filter using accumulated LiDAR point clouds, strictly grounding the extraction in physical scene geometry by removing unconstrained floaters.

For point-level LiDAR segmentation, we avoid naive nearest-neighbor assignments by employing an inverse-distance weighted voting scheme among the local Gaussian neighborhood. This spatial consensus out-votes isolated outliers and provides a native uncertainty metric via Shannon Entropy. Similarly, our continuous formulation enables resolution-agnostic semantic occupancy extraction. A target voxel is classified as occupied if it falls within a structurally filtered Gaussian. Its semantic class is derived from the mean consensus of the continuous semantic field queried at its eight bounding vertices, elegantly resolving boundary ambiguities while maintaining spatial coherence.

\subsection{Pseudo-Labelling as Selective Classification}
Existing pseudo-labelling baselines frequently evaluate performance strictly on the subset of points they label~\cite{gebraad_leap_2025, ghilotti_unilips_2026}. This evaluation protocol fails to provide a holistic understanding of a pipeline's efficacy, as it ignores the widely varying and arbitrary recall levels across different methods. 

To resolve this, we propose formulating 3D pseudo-labelling as a selective classification task~\cite{el-yaniv_foundations_2010}. By treating unassigned pseudo-labels as ``rejected'' predictions, we can evaluate disparate pipelines using robust risk-coverage trade-offs. We utilize Generalized Risk, which quantifies the joint probability that a model accepts a point for pseudo-labelling and that the assigned label is incorrect~\cite{traub_overcoming_2024}. This acts as a direct, practical measure of the risk of injecting noisy labels into downstream training. 

For a given class $k$ at a global coverage/recall threshold $c$, the selective risk is defined as $1 - \text{IoU}_{k,c}$, where IoU represents the Intersection over Union. Under severe class imbalance, standard selective global metrics often appear deceptively optimistic by hiding the disproportionate rejection of minority classes~\cite{saglam_selective_2026}. To ensure a fair, bias-aware evaluation across the entire dataset, we formulate our overall performance metric as the class-averaged Area Under the Generalized Risk-Coverage (AUGRC) curve:
\begin{equation}
\label{eq:augrc}
AUGRC = \int_{0}^{1} (1 - \text{mIoU}_c) \cdot c \, dc
\end{equation}
Bounded strictly between 0 and 0.5, a lower AUGRC indicates superior performance. By deriving this overall metric from the mean of the class-specific generalized risks, we provide a robust mathematical foundation to evaluate the true precision-recall reliability of any pseudo-labelling system.
\section{Experiments}
\label{sec:experiments}

\definecolor{cSkCar}{RGB}{100, 150, 245}
\definecolor{cSkBicycle}{RGB}{100, 230, 245}
\definecolor{cSkMotorcycle}{RGB}{30, 60, 150}
\definecolor{cSkTruck}{RGB}{80, 30, 180}
\definecolor{cSkOtherVehicle}{RGB}{0, 0, 255}
\definecolor{cSkPerson}{RGB}{255, 30, 30}
\definecolor{cSkBicyclist}{RGB}{255, 40, 200}
\definecolor{cSkMotorcyclist}{RGB}{150, 30, 90}
\definecolor{cSkRoad}{RGB}{255, 0, 255}
\definecolor{cSkParking}{RGB}{255, 150, 255}
\definecolor{cSkSidewalk}{RGB}{75, 0, 75}
\definecolor{cSkBuilding}{RGB}{255, 200, 0}
\definecolor{cSkFence}{RGB}{255, 120, 50}
\definecolor{cSkVegetation}{RGB}{0, 175, 0}
\definecolor{cSkTrunk}{RGB}{135, 60, 0}
\definecolor{cSkTerrain}{RGB}{150, 240, 80}
\definecolor{cSkPole}{RGB}{255, 240, 150}
\definecolor{cSkTrafficSign}{RGB}{255, 0, 0}

\begin{table*}[t]
\centering
\footnotesize
\caption{\textbf{Detailed Semantic Occupancy Performance on the SemanticKITTI validation set~\cite{behley_semantickitti_2019}.}
Following the evaluation protocol introduced by ~\cite{zhou_autoocc_2025}, the evaluated baselines are trained exclusively on Occ3D-nuScenes~\cite{tian_occ3d_2023} and tested in a zero-shot setting on SemanticKITTI~\cite{behley_semantickitti_2019} to assess their cross-dataset transferability. 
In contrast, AutoOcc~\cite{zhou_autoocc_2025} operates as an open-ended label generation pipeline annotating the target dataset directly. Compared against both zero-shot transferred baselines and AutoOcc~\cite{zhou_autoocc_2025} under this protocol, our method demonstrates superior performance. Input modalities are denoted as C (Camera) and L (LiDAR), with C+L indicating the use of both. Best results are highlighted in \textbf{bold} and second-best are \underline{underlined}.}
\label{tab:supp_semantickitti_comparison}
\renewcommand{\arraystretch}{1.2}
\setlength{\tabcolsep}{1.8pt}
\begin{tabular}{l | c | c c | c c c c c c c c c c c c c c c c c c}
\hline
\textbf{Method} & 
\textbf{Input} & 
\rotatebox{90}{Geo. IoU \% $\uparrow$\hspace{3pt}} & 
\rotatebox{90}{Sem. mIoU \% $\uparrow$\hspace{3pt}} & 
\rotatebox{90}{\textcolor{cSkCar}{\rule{1.2ex}{1.2ex}}\hspace{3pt}Car\hspace{3pt}} & 
\rotatebox{90}{\textcolor{cSkBicycle}{\rule{1.2ex}{1.2ex}}\hspace{3pt}Bicycle\hspace{3pt}} & 
\rotatebox{90}{\textcolor{cSkMotorcycle}{\rule{1.2ex}{1.2ex}}\hspace{3pt}Motorcycle\hspace{3pt}} & 
\rotatebox{90}{\textcolor{cSkTruck}{\rule{1.2ex}{1.2ex}}\hspace{3pt}Truck\hspace{3pt}} & 
\rotatebox{90}{\textcolor{cSkOtherVehicle}{\rule{1.2ex}{1.2ex}}\hspace{3pt}Oth.-vehicle\hspace{3pt}} & 
\rotatebox{90}{\textcolor{cSkPerson}{\rule{1.2ex}{1.2ex}}\hspace{3pt}Person\hspace{3pt}} & 
\rotatebox{90}{\textcolor{cSkBicyclist}{\rule{1.2ex}{1.2ex}}\hspace{3pt}Bicyclist\hspace{3pt}} & 
\rotatebox{90}{\textcolor{cSkMotorcyclist}{\rule{1.2ex}{1.2ex}}\hspace{3pt}Motorcyclist\hspace{3pt}} & 
\rotatebox{90}{\textcolor{cSkRoad}{\rule{1.2ex}{1.2ex}}\hspace{3pt}Road\hspace{3pt}} & 
\rotatebox{90}{\textcolor{cSkParking}{\rule{1.2ex}{1.2ex}}\hspace{3pt}Parking\hspace{3pt}} & 
\rotatebox{90}{\textcolor{cSkSidewalk}{\rule{1.2ex}{1.2ex}}\hspace{3pt}Sidewalk\hspace{3pt}} & 
\rotatebox{90}{\textcolor{cSkBuilding}{\rule{1.2ex}{1.2ex}}\hspace{3pt}Building\hspace{3pt}} & 
\rotatebox{90}{\textcolor{cSkFence}{\rule{1.2ex}{1.2ex}}\hspace{3pt}Fence\hspace{3pt}} & 
\rotatebox{90}{\textcolor{cSkVegetation}{\rule{1.2ex}{1.2ex}}\hspace{3pt}Vegetation\hspace{3pt}} & 
\rotatebox{90}{\textcolor{cSkTrunk}{\rule{1.2ex}{1.2ex}}\hspace{3pt}Trunk\hspace{3pt}} & 
\rotatebox{90}{\textcolor{cSkTerrain}{\rule{1.2ex}{1.2ex}}\hspace{3pt}Terrain\hspace{3pt}} & 
\rotatebox{90}{\textcolor{cSkPole}{\rule{1.2ex}{1.2ex}}\hspace{3pt}Pole\hspace{3pt}} & 
\rotatebox{90}{\textcolor{cSkTrafficSign}{\rule{1.2ex}{1.2ex}}\hspace{3pt}Traffic-sign\hspace{3pt}} \\
\hline
\small
GaussianOcc~\cite{gan_gaussianocc_2025} & C & 22.42 & 4.18 & 7.10 & 1.33 & 3.06 & 3.42 & 2.81 & 2.91 & 0.00 & 0.00 & 15.80 & 0.00 & 10.43 & 2.55 & 0.00 & 22.11 & 0.00 & 3.78 & 0.00 & 0.00 \\
OVO~\cite{tan_ovo_2023} & C & 20.94 & 5.83 & 12.70 & 0.40 & 0.20 & 0.70 & 3.50 & 0.74 & 0.90 & 0.00 & 19.44 & \underline{0.68} & 24.81 & 11.70 & 3.50 & 15.62 & 2.31 & 4.86 & 0.60 & 2.20 \\
SurroundOcc~\cite{wei_surroundocc_2023} & L & 27.83 & 6.39 & 23.19 & 1.52 & 6.71 & 3.16 & \underline{4.81} & 4.37 & 0.00 & 0.00 & 24.32 & 0.00 & 11.98 & 5.79 & 0.00 & 19.14 & 0.00 & \textbf{9.95} & 0.00 & 0.00 \\
VLM-LiDAR~\cite{zhou_autoocc_2025} & C+L & 28.12 & 5.32 & 19.17 & 2.04 & 2.13 & 5.89 & 3.31 & 2.64 & 0.00 & 0.00 & 19.02 & 0.00 & 16.58 & 3.59 & 0.00 & 14.98 & 0.00 & 6.31 & 0.00 & 0.00 \\
AutoOcc-V~\cite{zhou_autoocc_2025} & C & 35.64 & 9.36 & 22.29 & 4.71 & \underline{10.35} & \underline{8.78} & 3.89 & 7.54 & \textbf{1.38} & \underline{3.60} & 26.14 & 0.59 & 15.66 & 4.14 & 4.34 & 18.87 & 5.36 & \underline{9.84} & \underline{14.32} & 6.62 \\
AutoOcc-M~\cite{zhou_autoocc_2025} & C+L & \underline{41.23} & \underline{12.76} & \underline{24.60} & \underline{7.83} & 9.30 & 8.39 & \textbf{4.92} & \textbf{11.18} & \underline{1.27} & \textbf{5.23} & \underline{44.74} & 0.33 & \underline{24.43} & \underline{17.01} & \underline{5.71} & \underline{29.12} & \underline{5.97} & 5.85 & \textbf{15.17} & \underline{8.72} \\
\textbf{SplatLabel (Ours)} & C+L & \textbf{47.36} & \textbf{15.56} & \textbf{28.04} & \textbf{10.91} & \textbf{16.74} & \textbf{9.65} & 0.28 & \underline{7.65} & 0.26 & 0.01 & \textbf{48.58} & \textbf{11.30} & \textbf{30.94} & \textbf{22.58} & \textbf{11.79} & \textbf{32.96} & \textbf{19.15} & 7.37 & 12.10 & \textbf{9.71} \\
\hline
\end{tabular}
\end{table*}

To validate the efficacy of our proposed representation, we evaluate its performance on two downstream tasks: 3D LiDAR segmentation and semantic occupancy prediction. 

\subsection{Experimental Setup}
We evaluate our method on the SemanticKITTI~\cite{behley_semantickitti_2019} validation set (Sequence 08). Semantic features are extracted zero-shot using the SAM 3~\cite{carion_sam_2026} foundation model. Crucially, we eschew complex prompt engineering, opting instead to directly prompt the dataset class names to evaluate the native robustness of our differentiable learning backend. Due to the inherent VRAM constraints associated with optimizing dense 3D Gaussian Splatting over long durations, we divide the sequence into independent, approximately 20-second continuous scenes. These localized scenes are trained independently on NVIDIA A40 GPUs and subsequently stitched together using a primitive alignment strategy during inference.

\subsection{LiDAR Segmentation and Recall Evaluation}
To ensure fair comparisons with existing baselines, we evaluate our results using the 11 common merged classes (car, bicycle, motorcycle, other-vehicle, person, road, sidewalk, other-ground, manmade, vegetation, and terrain) and the six  categories (flat, construction, object, nature, human, and vehicle) defined by the KITTI-360 benchmark~\cite{liao_kitti-360_2023}. Qualitative segmentation results are visualized in the top rows of Fig. \ref{fig:qualitative_seg}.

Through direct correspondence with the respective authors, we determined that LeAP~\cite{gebraad_leap_2025} generates labels for 65.3\% of the points, while UniLiPs~\cite{ghilotti_unilips_2026} labels an average of 47.32\% of the points on the training set. Because our methodology inherently supports continuous evaluation across all recall thresholds via our selective classification formulation, we benchmark our pipeline directly against these baselines at their exact recall levels to ensure absolute fairness. As illustrated in Figure \ref{fig:coverage_tradeoff}, SplatLabel maintains a superior performance profile across all recall levels, consistently outperforming baseline operating points even when accounting for variance. Global quantitative results at these specific recall thresholds are summarized in Table \ref{tab:supp_comparison}. As detailed in Table \ref{tab:supp_comparison}, our method struggles specifically with the `terrain' class. This is an expected consequence of our zero-shot VFM distillation without domain-specific prompt engineering, as 2D VFM could conflate terrain with vegetation. However, because both classes belong to the same broader category, our category-level mIoU (cat. mIoU) natively captures this structural understanding without penalty.

\subsection{Semantic Occupancy Prediction}
Beyond point-level segmentation, we evaluate our performance on semantic occupancy prediction against AutoOcc~\cite{zhou_autoocc_2025}, the current state-of-the-art pseudo-labelling method. Following their evaluation protocol, we assess performance across 18 semantic classes, excluding the ambiguous ``other flat/ground" class to ensure a fair comparison~\cite{zhou_autoocc_2025}. As qualitatively demonstrated in the bottom rows of Fig. \ref{fig:qualitative_seg}, our explicit kinematic parameterization successfully prevents the temporal smearing artifacts often present in aggregated ground truth data.

As shown in Table \ref{tab:supp_semantickitti_comparison}, our method demonstrates superior Geometric and Semantic IoU. To ensure the statistical significance of these improvements, we evaluated our method across five random seeds. We observed a minimal standard deviation of just $0.066$ and $0.041$ percentage points for Geometric IoU and Semantic mIoU respectively. These microscopic variances are orders of magnitude smaller than the absolute performance margins we achieve over previous baselines, thoroughly solidifying the stability and superiority of our approach.

\subsection{Spatial Domain and Generalized Risk Evaluation}
To establish standardized baselines for pseudo-label reliability under varying geometric constraints, we evaluate the Area Under the Generalized Risk-Coverage curve (AUGRC) across three distinct spatial visibility subsets (see Table \ref{tab:augrc_all}):

\begin{itemize}
    \item \textit{All Points (100\% recall):} The complete 360-degree LiDAR frame, evaluated irrespective of camera overlap.
    
    \item \textit{Camera (92.0\% recall):} Points projecting into any camera frustum across the temporal sequence, inclusive of physically occluded points.
    
    \item \textit{Unoccluded (62.3\% recall):} Points within the frustum that are physically unobstructed. We determine this by comparing a LiDAR point's true depth to the rendered depth map; a point is classified as unoccluded if its true depth does not exceed the rendered depth by more than a 0.5-meter tolerance.
\end{itemize}

Lower AUGRC values indicate a more favorable accuracy-coverage trade-off, demonstrating our method's robustness across diverse semantic classes without restricting evaluation to heavily curated domains. \looseness = -1

% Color definitions for all tables
\definecolor{cCar}{RGB}{100, 150, 245}
\definecolor{cBicycle}{RGB}{100, 230, 245}
\definecolor{cMotorcycle}{RGB}{30, 60, 150}
\definecolor{cOtherVehicle}{RGB}{0, 0, 255}
\definecolor{cPerson}{RGB}{255, 30, 30}
\definecolor{cRoad}{RGB}{255, 0, 255}
\definecolor{cSidewalk}{RGB}{75, 0, 75}
\definecolor{cOtherGround}{RGB}{175, 0, 75}
\definecolor{cManmade}{RGB}{255, 200, 0}
\definecolor{cVegetation}{RGB}{0, 175, 0}
\definecolor{cTerrain}{RGB}{150, 240, 80}
\definecolor{cPole}{RGB}{255, 240, 150}

\begin{table*}[t]
\centering
\caption{\textbf{Detailed Spatial Domain Evaluation.} Per-class and per-category AUGRC evaluated across varying spatial visibility subsets on the SemanticKITTI validation set~\cite{behley_semantickitti_2019}. Lower values indicate better reliability.}
\label{tab:augrc_all}
\renewcommand{\arraystretch}{1.2} 
\setlength{\tabcolsep}{3pt} 
\begin{tabular}{l | c | c | c c c c c c c c c c c | c | c c c c c c}
\hline
\textbf{Eval. Domain} & 
\rotatebox{90}{Recall \%\hspace{4pt}} & 
\rotatebox{90}{AUGRC \% $\downarrow$\hspace{4pt}} & 
\rotatebox{90}{\textcolor{cCar}{\rule{1.5ex}{1.5ex}}\hspace{3pt}Car\hspace{4pt}} & 
\rotatebox{90}{\textcolor{cBicycle}{\rule{1.5ex}{1.5ex}}\hspace{3pt}Bicycle\hspace{4pt}} & 
\rotatebox{90}{\textcolor{cMotorcycle}{\rule{1.5ex}{1.5ex}}\hspace{3pt}Motorcycle\hspace{4pt}} & 
\rotatebox{90}{\textcolor{cOtherVehicle}{\rule{1.5ex}{1.5ex}}\hspace{3pt}Oth.-vehicle\hspace{4pt}} & 
\rotatebox{90}{\textcolor{cPerson}{\rule{1.5ex}{1.5ex}}\hspace{3pt}Person\hspace{4pt}} & 
\rotatebox{90}{\textcolor{cRoad}{\rule{1.5ex}{1.5ex}}\hspace{3pt}Road\hspace{4pt}} & 
\rotatebox{90}{\textcolor{cSidewalk}{\rule{1.5ex}{1.5ex}}\hspace{3pt}Sidewalk\hspace{4pt}} & 
\rotatebox{90}{\textcolor{cOtherGround}{\rule{1.5ex}{1.5ex}}\hspace{3pt}Oth.-ground\hspace{4pt}} & 
\rotatebox{90}{\textcolor{cManmade}{\rule{1.5ex}{1.5ex}}\hspace{3pt}Manmade\hspace{4pt}} & 
\rotatebox{90}{\textcolor{cVegetation}{\rule{1.5ex}{1.5ex}}\hspace{3pt}Vegetation\hspace{4pt}} & 
\rotatebox{90}{\textcolor{cTerrain}{\rule{1.5ex}{1.5ex}}\hspace{3pt}Terrain\hspace{4pt}} & 
\rotatebox{90}{cat. AUGRC \% $\downarrow$\hspace{4pt}} & 
\rotatebox{90}{\textcolor{cRoad}{\rule{1.5ex}{1.5ex}}\hspace{3pt}Flat\hspace{4pt}} & 
\rotatebox{90}{\textcolor{cManmade}{\rule{1.5ex}{1.5ex}}\hspace{3pt}Construction\hspace{4pt}} & 
\rotatebox{90}{\textcolor{cPole}{\rule{1.5ex}{1.5ex}}\hspace{3pt}Object\hspace{4pt}} & 
\rotatebox{90}{\textcolor{cVegetation}{\rule{1.5ex}{1.5ex}}\hspace{3pt}Nature\hspace{4pt}} & 
\rotatebox{90}{\textcolor{cPerson}{\rule{1.5ex}{1.5ex}}\hspace{3pt}Human\hspace{4pt}} & 
\rotatebox{90}{\textcolor{cCar}{\rule{1.5ex}{1.5ex}}\hspace{3pt}Vehicle\hspace{4pt}} \\
\hline
All Points & 100 & 19.4 & 10.2 & 27.2 & 17.3 & 10.1 & 17.0 & 3.4 & 10.2 & 46.5 & 11.0 & 14.4 & 45.9 & 11.3 & 4.2 & 11.2 & 17.9 & 5.9 & 18.5 & 10.4 \\
Camera & 92.0 & 19.2 & 9.8 & 27.5 & 19.5 & 9.8 & 14.7 & 3.1 & 9.3 & 46.4 & 10.7 & 14.2 & 45.9 & 10.1 & 3.7 & 10.8 & 16.0 & 5.5 & 14.7 & 9.9 \\
Unoccluded & 62.3 & 16.1 & 5.2 & 23.3 & 15.1 & 5.4 & 10.3 & 1.9 & 5.9 & 43.7 & 7.2 & 13.3 & 45.5 & 6.4 & 1.9 & 6.6 & 12.5 & 3.4 & 8.8 & 5.1 \\
\hline
\end{tabular}
\end{table*}

\subsection{Computational Resources}
All experiments are executed on a cluster utilizing 4 NVIDIA A40 GPUs concurrently. Because our pipeline divides the dataset into independent, approximately 20-second continuous scene splits, the optimization process is highly parallelizable, requiring a total end-to-end training time of 15.28 GPU hours for 3D LiDAR segmentation and 33.67 GPU hours for semantic occupancy prediction on the complete SemanticKITTI validation set. During the inference phase, the label generation process operates efficiently, taking an average of 0.28 and 4.62 seconds per frame for LiDAR segmentation and semantic occupancy prediction respectively, as measured on a single NVIDIA A40 GPU.
\section{Ablation Study}
\label{sec:ablations}

To isolate the individual contributions of our pipeline's spatial, kinematic, and temporal mechanisms, we perform a leave-one-out ablation study, summarized in Table \ref{tab:ablation_study}. \looseness = -1

\subsubsection{Semantic and Geometric Priors} 
Replacing our continuous soft labels with traditional hard semantic assignments (w/ Hard Semantic Labels) severely degrades overall performance. This validates that backpropagating continuous soft semantics is critical for natively resolving categorical ambiguities through spatiotemporal consensus. Geometric priors are equally vital; removing Virtual Depth Maps, which integrate 360-degree LiDAR into unobserved regions, causes a massive drop in Geometric IoU. Finally, bypassing Geometry Filtering allows unconstrained floating artifacts and 2D-optimized noise to corrupt 3D pseudo-label generation, uniformly harming all metrics.

\subsubsection{Spatiotemporal Kinematics} 
Explicit kinematic parameterization is essential for accurately separating dynamic actors from static environments. Disabling the Gated Deformation Mask allows static structures to improperly track micro-motions rather than remaining firmly clamped to zero, introducing widespread geometric jitter that sharply reduces geometric and semantic fidelity. Furthermore, decoupling the lifespan from the deformation mask ($-$ Coupled Lifespan) causes a consistent drop across all metrics. This indicates that ensuring static primitives retain a full, sequence-wide footprint is vital for maintaining global scene stability.

\begin{table}[ht]
\centering
\caption{\textbf{Ablation Study.} Contribution of key pipeline components. 
The $(-)$ denotes the removal of a proposed component, while $(\text{w/})$ denotes the substitution of a baseline methodology. Values represent the performance delta relative to the Full Pipeline. \textcolor{ForestGreen}{Green} values indicate that removing the component degrades overall performance (a positive delta for AUGRC or a negative delta for IoU), thereby validating its necessity.}
\label{tab:ablation_study}
\renewcommand{\arraystretch}{1.2}
\setlength{\tabcolsep}{3pt}
\begin{tabular}{l | c | c | c}
\hline
\textbf{Ablation Variant} & 
\footnotesize \textbf{AUGRC} & 
\footnotesize \textbf{Geo. IoU} & 
\footnotesize \textbf{Sem. mIoU} \\
\hline
\textbf{SplatLabel: Full Pipeline} & \textbf{19.4} & \textbf{47.4} & \textbf{15.6} \\
w/ Hard Semantic Labels & \textcolor{ForestGreen}{+6.5} & \textcolor{ForestGreen}{-1.9} & \textcolor{ForestGreen}{-1.2} \\
$-$ Virtual Depth Maps & \textcolor{red}{-0.5} & \textcolor{ForestGreen}{-15.7} & \textcolor{ForestGreen}{-3.4} \\
$-$ Gated Deformation Mask & \textcolor{ForestGreen}{+2.0} & \textcolor{ForestGreen}{-12.7} & \textcolor{ForestGreen}{-4.3} \\
$-$ Coupled Lifespan & \textcolor{ForestGreen}{+1.1} & \textcolor{ForestGreen}{-5.9} & \textcolor{ForestGreen}{-1.5} \\
$-$ Geometry Filtering & \textcolor{ForestGreen}{+3.7} & \textcolor{ForestGreen}{-3.9} & \textcolor{ForestGreen}{-3.2} \\
\hline
\end{tabular}
\end{table}
\section{Conclusion}
\label{sec:conclusion}

We presented SplatLabel, an automated 4D Gaussian Splatting pipeline designed specifically for 3D semantic pseudo-labelling and volumetric occupancy prediction. At its core, SplatLabel overcomes the bottlenecks of dynamic scene reconstruction by introducing an explicit temporal manifold. By parameterizing continuous motion, temporal lifespans, and physical existence boundaries as intrinsic primitive properties, our framework successfully tracks dynamic actors while cleanly preserving static backgrounds. This approach entirely eliminates the need for pre-annotated 3D bounding boxes. \looseness = -1

To robustly ground this temporal representation, we utilize a Spatiotemporal Sampling-Based Density Controller built on ~\cite{kheradmand_3d_2024} that directly distills continuous soft semantics from 2D foundation models~\cite{carion_sam_2026} and injects 360-degree LiDAR priors via virtual depth maps. Furthermore, to address systemic flaws in standard pseudo-label evaluation, we reformulated the task as selective classification using the AUGRC~\cite{traub_overcoming_2024} metric, providing a mathematically grounded, bias-aware measure of precision-recall reliability. Experiments on SemanticKITTI demonstrate that SplatLabel consistently outperforms state-of-the-art baselines across arbitrary recall levels, establishing a robust, annotation-free foundation for dynamic scene understanding.\looseness = -1

\subsection{Limitations}
Our approach is constrained by the high VRAM requirements inherent to dense 3DGS~\cite{kerbl_3d_2023} optimization and relies on precise camera-LiDAR synchronization. Additionally, because the pipeline depends heavily on camera imagery for dense semantics and LiDAR for structural geometry, its robustness inherently degrades in adverse lighting or weather conditions where sensor inputs are compromised. 

\subsection{Future Work}
For future work, we plan to validate the practical efficacy of our generated labels directly on downstream tasks. This will involve training downstream semantic segmentation and occupancy models purely on our generated annotations and benchmarking their final performance against models trained on human-annotated ground truth. Finally, because our pipeline natively outputs predictive confidence scores via spatial consensus, future research will explore integrating these uncertainty metrics directly into downstream training processes to dynamically weight loss functions.

\bibliographystyle{IEEEtran}
\bibliography{IEEEabrv, references}

\end{document}